\documentclass[12pt]{iopart}
\usepackage{iopams}
\usepackage{amsfonts}
\usepackage{amsmath}
\usepackage{amssymb}
\usepackage{mathabx}
\usepackage{mathrsfs}
\usepackage{tensor}
\usepackage{esint}
\usepackage{tikz}
\usepackage{multirow}
\usepackage{harvard}
\DeclareMathAlphabet{\mathpzc}{OT1}{pzc}{m}{it}
\def\R{{\Bbb R}}

\def\ep{\epsilon}

\def\f{\frac}

\def\x{\mathbf{x}}
\def\y{\mathbf{y}}

\def\Ndata{\mathfrak{n}_{\mbox{\tiny data}}}

\def\etal{{\it et al.}}

\begin{document}

\title[Framelet-Aided Deep Learning Network]{Framelet Pooling Aided Deep Learning Network : The Method to Process High Dimensional Medical Data}

\author{Chang Min Hyun\dag, Kang Cheol Kim\dag, Hyun Cheol Cho\dag, Jae Kyu Choi\ddag\footnote[5]{To whom correspondence should be addressed (jaycjk@tongji.edu.cn)} $\;$ and Jin Keun Seo\dag}
\address{\dag Department of Computational Science and Engineering, Yonsei University, Seoul, Korea}
\address{\ddag School of Mathematical Sciences, Tongji University, Shanghai, 200092, China}

\begin{abstract}
Machine learning-based analysis of medical images often faces several hurdles, such as the lack of training data, the curse of dimensionality problem, and the generalization issues. One of the main difficulties is that there exists computational cost problem in dealing with input data of large size matrices which represent medical images. The purpose of this paper is to introduce a framelet-pooling aided deep learning method for mitigating computational bundle, caused by large dimensionality. By transforming high dimensional data into low dimensional components by filter banks with preserving detailed information, the proposed method aims to reduce the complexity of the neural network and computational costs significantly during the learning process. Various experiments show that our method is comparable to the standard unreduced learning method, while reducing computational burdens by decomposing large-sized learning tasks into several small-scale learning tasks.
\end{abstract}

\maketitle

\section{Introduction}
Recently, medical imaging is experiencing a paradigm shift due to a remarkable and rapid advance in deep learning techniques. Deep learning techniques have expanded our ability by sophisticated “disentangled representation learning” through training data, and appear to show superiority of performance in various medical imaging problems including undersampled magnetic resonance imaging(MRI), sparse-view computed tomography(CT), artifact reduction, organ segmentation, and automated disease detection. In particular, U-net \cite{Ronneberger2015}, a kind of convolutional neural network, seems to show remarkable capability of learning image representations. However, there are some hurdles to overcome, one of which comes from the high dimensionality, i.e. the high resolution or the large size, of medical images. This paper addresses a way to resolve this issue through a so-called \textit{framelet pooling aided deep learning network}.

Machine learning performance is closely related to the number, the quality, and the pixel dimensionality of the sampled data. For ease of explanation, let us consider a simple question to learn an unknown function $f:[0,1]^d\mapsto[0,1]$ from a given sample $(\x,y)$, where $\x$ is an input gray scale image lying in $[0,1]^d$ and $y=f(\x)$ is the corresponding output on the interval $[0,1]$. Then one can ask how many training samples are needed to approximate $f$ with a given tolerance $\epsilon >0$. It is well-known that for Lipschitz continuous function $f$, we need to sample $O(\ep^{-d})$ points \cite{mallat2016}. In addition, the author in \cite{barron1994} observed that the estimation error of the function $f$ by 1 hidden layer neural networks is given by $O(\f{c_f}{\mathfrak{m}}) +O\left(\f{\mathfrak{m}d}{\Ndata}\log \Ndata \right)$, where $\Ndata$ is the number of training data, $\mathfrak{m}$ is the number of neurons in the hidden layer, and $c_f$ is a constant depending on the regularity of $f$. This means that in the case of $d=512^2$ (i.e. considering $512\times512$ images) and $\mathfrak{m}=d$, we roughly need huge training data $\Ndata=O(10^{12})$ to achieve the error of $O(10^{-1})$. This high number of required training data makes the problem intractable, especially when data lies in the high dimensional space. Such a phenomenon is referred as the \emph{curse-of-dimensionality} in approximation sense. Even though the effect of dimensionality on deep networks is relatively weaker than shallow ones \cite{bruna2013,pascanu2013,brainMIT2016} in approximation sense, deep learning requires huge computational scale for training process. Thus, deep networks with high dimensional data also experience the curse-of-dimensionality in terms of computational burden.

In the literature, framelets are known to be effective in capturing key information of images. This is due to the multiscale structure of the framelet systems, and the presence of both low pass and high pass filters in the filter banks, which are desirable in sparsely approximating images without loss of information \cite{bin2017}. In this work, we propose a framelet-based deep learning method to reduce computational burdens for dealing with high dimensional data in the learning process. This method, called a \textit{framelet pooling}, is based on the decomposition of a $d$-dimensional input-output pair $(\x,\y)$ into several $d/2^{2k}$-dimensional pairs $\{ (\mathpzc{W}_{k,\alpha}\x,\mathcal{W}_{k,\alpha}\y):\alpha=1,\cdots,r\}$, where each $\mathpzc{W}_{k,\alpha}$ and $\mathcal{W}_{k,\alpha}$ are $d/2^{2k}\times d$ matrices corresponding to $k$th level framelet packet transform \cite{mallat2009}. Instead of learning the pair of high dimensional original data $(\x,\y)$, the proposed method tries to learn much lower dimensional pairs $(\mathcal{W}_{k,\alpha}\x,\mathpzc{W}_{k,\alpha}\y)$ in parallel passion, so that we can achieve the computational efficiency in dealing with the large size images.

As an application of our proposed method, we deal with the undersampled MRI \cite{Hyun2018} and the sparse-view CT problem \cite{jin2017}, where huge memory problems may arise in recovering high resolution images. Experiments on undersampled MRI and sparse-view CT show that our framelet pooling aided reduced method provides very similar performance to the standard unreduced method, while reducing the computation time greatly by reducing the dimension of inputs and learning parameters in neural networks.


\section{Method} \label{Method}
Both undersampled MRI and sparse-view CT problem aim to find a reconstruction function $f$, which maps from an undersampled data $\mathbf{P}^{\sharp}$ (violating Nyquist criteria) to a clinically meaningful tomographic image $\y$. Here, the undersampled data  $\textbf{P}^{\sharp}$ can be expressed as the subsampling of the fully-sampled data $\mathbf{P}$ (satisfying the Nyquist criterion),
\begin{equation}
\textbf{P}^{\sharp} = \mathcal{S} \textbf{P}
\end{equation}
where $\mathcal{S}$ is a subsampling operator. The standard MRI and CT use the fully-sampled data $\mathbf{P}$ to provide tomographic images, where the reconstruction functions $f$ in MRI and CT are the inverse Fourier transform and inverse Radon transform, respectively. However, when we use the undersampled data $\textbf{P}^{\sharp}$, these standard methods do not work as the Nyquist criterion is not satisfied any more. (See  Fig. \ref{Fig1} and Fig. \ref{Fig2}.) For the sake of clarity, we shall briefly state the mathematical framework of undersampled MRI and sparse-view CT in the following subsections.

\subsection{Undersampled MRI}
\begin{figure*}[h!]
	\centering
	\includegraphics[width=0.8\textwidth]{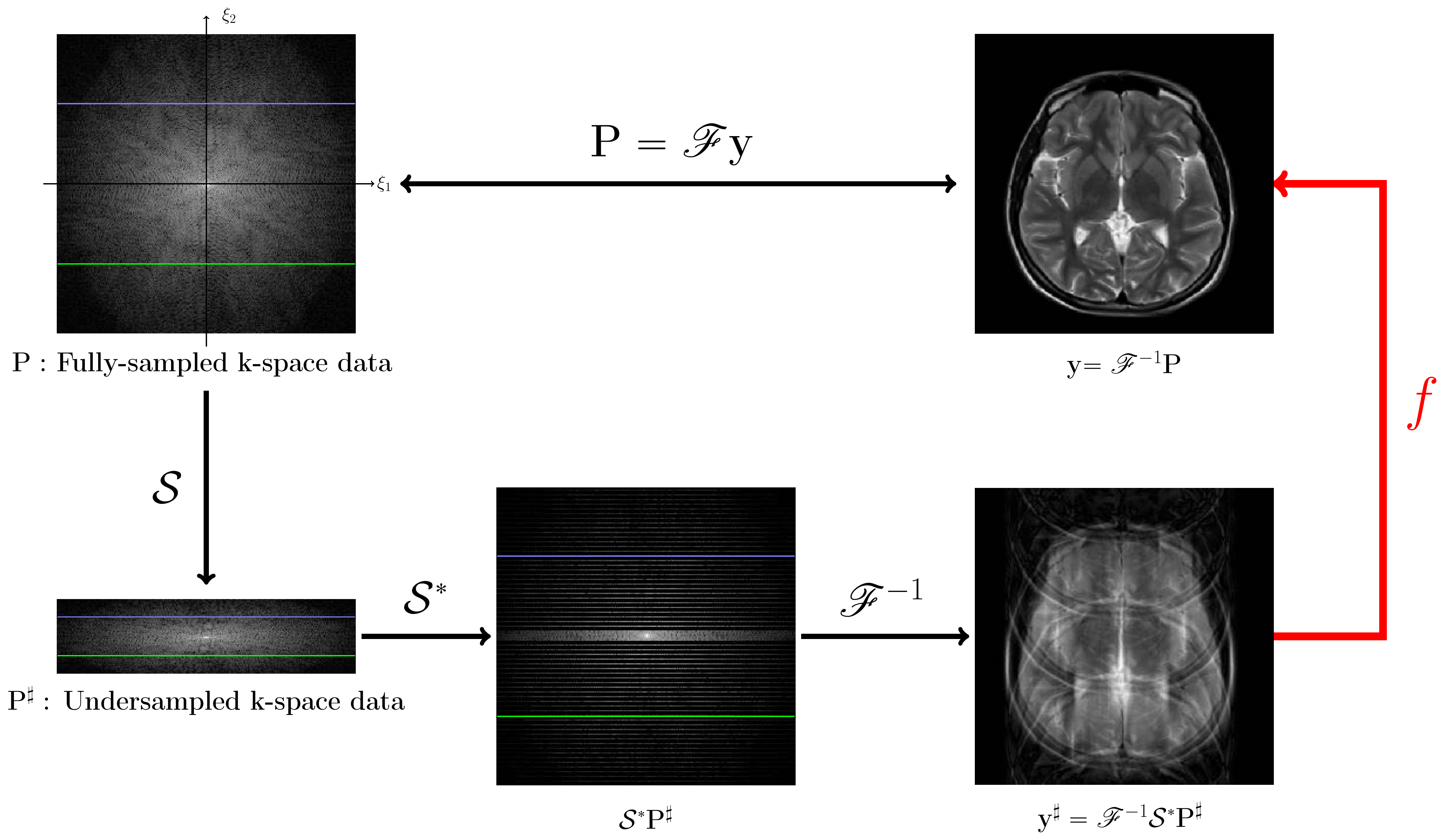}
	\caption{A $k$-space data in the standard MRI system is often measured with respect to two encoding directions, frequency direction $\xi_1$ and phase encoding direction $\xi_2$. A total time for MR imaging is, roughly speaking, is proportional to the number of phase encodings, since each phase encoding requires an repeated excitation process of nuclear spin. Thus, the undersampled MRI reconstruction problem deal with how to reduce the number of phase encoding lines by undersampling $\textbf{P}$ such like $\textbf{P}^{\sharp}$, and keep its image quality simultaneously. However, the reconstruction image $\y^{\sharp}$ obtained from the direct inversion method contains the aliasing artifact in the reconstruction domain, according to the Poisson summation formula. Therefore, we aims to develop a function $f$ to recover the artifacted image $\y^{\#}$ to the high quality image $\y$.}
	\label{Fig1}
\end{figure*}

Let $\y(z)$ be a distribution of nuclear spin density at the position $z=(z_1,z_2)$. The measured k-space data $\textbf{P}$ is governed by the Fourier relation,
\begin{equation} \label{ForwardCT}
\textbf{P} = \mathscr{F}\y = \int_{\mathbb{R}^2} \y(z) e^{-2\pi i z\cdot \xi} dz
\end{equation}
where $\xi = (\xi_1,\xi_2)$ \cite{Nishimura2010}. Therefore, with the fully-sampled data $\textbf{P}$, the reconstruction image $\y$ can be obtained by taking the inverse Fourier transform to the measured data $\textbf{P}$,
\begin{equation} \label{InverseMRI}
\y = \mathscr{F}^{-1} \textbf{P}
\end{equation}
Note that the direct inversion method \eqref{InverseMRI} can also be applied to the undersampled data $\textbf{P}^{\sharp}$,
\begin{equation}\label{underMRI}
\y^{\sharp} = \mathscr{F}^{-1} \mathcal{S}^* \textbf{P}^{\sharp}
\end{equation}
Here, $\mathcal{S}^*$ is an adjoint operator of $\mathcal{S}$ in the $\ell^2$ space. However, the image $\y^{\sharp}$ obtained from \eqref{underMRI} contains aliasing artifacts as $\textbf{P}^{\sharp}$ violates the Nyquist criterion(See Fig. \ref{Fig1}).

\subsection{Sparse-view CT}
\begin{figure*}[h!]
	\centering
	\includegraphics[width=1\textwidth]{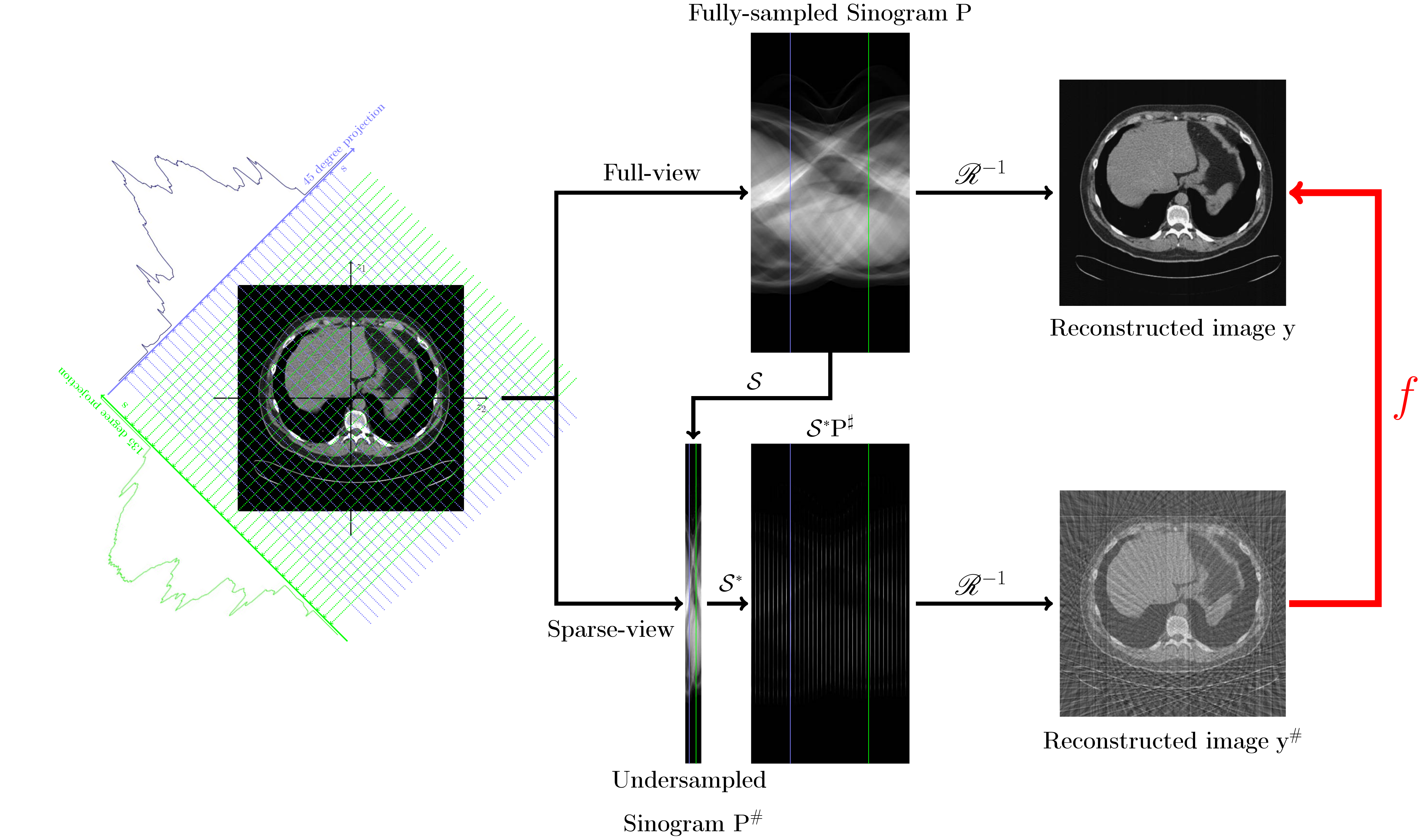}
	\caption{A sparse-view sinogram data is acquired from a sparse measurement of CT scanner with respect to projection view. The sparse-angle sinogram data can be considered as a subsampled sinogram  $\mathcal{S}\mbox{\textbf{P}}$ from a full-view sinogram $\mbox{\textbf{P}}$. The direct application of $\mathscr{R}$ to sparse-angle sinogram data with zero-filling $\mathcal{S}^*$ generates the streaking artifacts in the reconstruction image $\y^{\sharp}$. Main objective of sparse-angle computed tomography is to find a function $f$ to recover the artifacted image $\y^{\#}$ to the high quality image $\y$.}
	\label{Fig2}
\end{figure*}

In CT, the tomographic image $\y(z)$ can be regarded as the distribution of linear attenuation coefficients at the position $z=(z_1,z_2)$. For CT data acquisition, X-ray beams are transmitted at various directions $\theta:=(\cos\varphi,\sin\varphi),~~0\le \varphi\le 2\pi$. Under the assumption of monochromatic X-ray generation, the projection data $\textbf{P}$ at the direction $\theta$ is dictated by the following Radon transform
\begin{equation} \label{FowradCT}
\textbf{P} =  \mathscr{R}\y = \int_{L_{\theta,s}} \y(z) d\ell_z
\end{equation}
where $L_{\theta,s}$ is the projection line $L_{\theta,s}:= \{z\in \R^2~:~\theta\cdot z=s\}$ \cite{seo2013}. With the fully-sampled data $\textbf{P}$ satisfying the Nyquist criterion, $\y$ can be reconstructed by the inverse Radon transform
\begin{equation} \label{InverseCT}
\y = \mathscr{R}^{-1}\textbf{P}
\end{equation}

For the undersampled data $\textbf{P}^{\sharp}$, which is measured with the low sampling frequency along the projection-view, we can apply the direct inversion formula \eqref{InverseCT} by filling zeros to unmeasured parts of undersampled data
\begin{equation} \label{InverseCT2}
\y^{\sharp} = \mathscr{R}^{-1}\mathcal{S}^*\textbf{P}^{\sharp}
\end{equation}
However, the reconstruction image $\y^{\sharp}$ contains streaking artifacts, which result from the violation of Nyquist criterion. Fig. \ref{Fig2} shows the schematic and visual descriptions of the sparse-view CT problem.

\subsection{Main result: Undersampled reconstruction using framelet and deep learning} \label{MR}

The objective of the undersampled reconstruction problem is to develop a deartifacting map $f$, which converts $\y^{\sharp} \in \mathbb{R}^{d^2}$(artifacted image) to $\y \in \mathbb{R}^{d^2}$(artifact removed image) with $d^2$ being a pixel dimension of reconstructed image. In particular, deep learning techniques, such as U-net, infer $f$ by minimizing training data-fidelity :
\begin{equation} \label{directDL}
f = \underset{f \in \mathbb{D}\mathbb{L}_{\tiny \mbox{net}}}{\mbox{argmin}} \sum_{i=1}^{N} \mathscr{L} (f(\x^{(i)}),\y^{(i)})
\end{equation}
using a set of training data $(\x^{(i)}, \y^{(i)})_{i=1}^{N}$. Here, $N$ is the number of training data, $\x^{(i)}$ denotes the artifact image instead of  $(\y^{\sharp})^{(i)}$, $\mathbb{DL}_{\tiny \mbox{net}}$ is a set of all learnable functions from a user-defined deep learning network architecture, and $\mathscr{L}$ is a user-defined energy-loss function to evaluate the metric between deep learning output $f(\x^{(i)})$ and label $\y^{(i)}$. However, if the pixel dimension of input increases, the total computational complexity in the training process increase largely. To address this curse-of-dimensionality issue, we propose the framelet pooling aided deep learning method to learn the deartifacting map $f$ indirectly.

For sake of clarity, we first provide brief introduction to the framelet.
Let $\phi \in \mbox{L}^2(\mathbb{R}^{2})$ be a refinable function, which satisfies
\begin{equation} \label{SR}
\widehat{\phi}(\boldsymbol\xi) = \widehat{\textbf{q}_0}(2^{-1}\boldsymbol\xi)\widehat{\phi}(2^{-1}\boldsymbol\xi), \; \forall \boldsymbol\xi \in \mathbb{R}^2
\end{equation}
where $\widehat{\cdot}$ denotes the Fourier transform operator. Then, framelet functions $\Psi = \{ \psi_\alpha : 1\leq\alpha\leq r \} \subseteq \mbox{L}^2(\mathbb{R}^2)$ is generated by
\begin{equation} \label{SR2}
\widehat{\psi}_{\alpha}(\boldsymbol\xi) = \widehat{\textbf{q}_\alpha}(2^{-1}\boldsymbol\xi)\widehat{\phi}(2^{-1}\boldsymbol\xi), \; \forall \boldsymbol\xi \in [0,\pi]^2~\&~1 \leq \alpha \leq r
\end{equation}
where $\textbf{q}_\alpha$ satisfies the unitary extension principle \cite{ron1997,bin2012},
\begin{align} \label{UEP}
\sum_{\alpha=0}^{r} |\widehat{\textbf{q}_\alpha}(\boldsymbol\xi)|^2=1,	\sum_{\alpha=0}^{r} \widehat{\textbf{q}_\alpha}(\boldsymbol\xi)\overline{\widehat{\textbf{q}_\alpha}(\boldsymbol\xi+\boldsymbol\nu)}=0, \\ \notag \forall \boldsymbol\xi \in [0,\pi]^2, \forall \boldsymbol\nu \in \{0,\pi\}^2 \backslash \{\textbf{0}\}.
\end{align}
Then the corresponding affine system $X(\Psi)=\{\psi_{\alpha,n,\textbf{k}} = 2^{n}\psi_{\alpha}(2^n \cdot - \textbf{k}):1\leq\alpha\leq r, n \in \mathbb{Z}, \textbf{k}\in \mathbb{Z}^2 \}$ forms a tight frame for $\mbox{L}^2(\mathbb{R}^2)$, and the filter banks $\{ \textbf{q}_{\alpha} \}_{i \in \{ 0,1,\cdots,r \}}$ form a tight frame on $\ell^2(\mathbb{Z}^2)$. This means that these filter banks project high dimensional data into low dimensional space without any information loss. In other words, an invertible decomposition procedure, called framelet decomposition, can be defined from these filter banks.

\begin{figure*}[h!]
	\centering
	\includegraphics[width=0.8\textwidth]{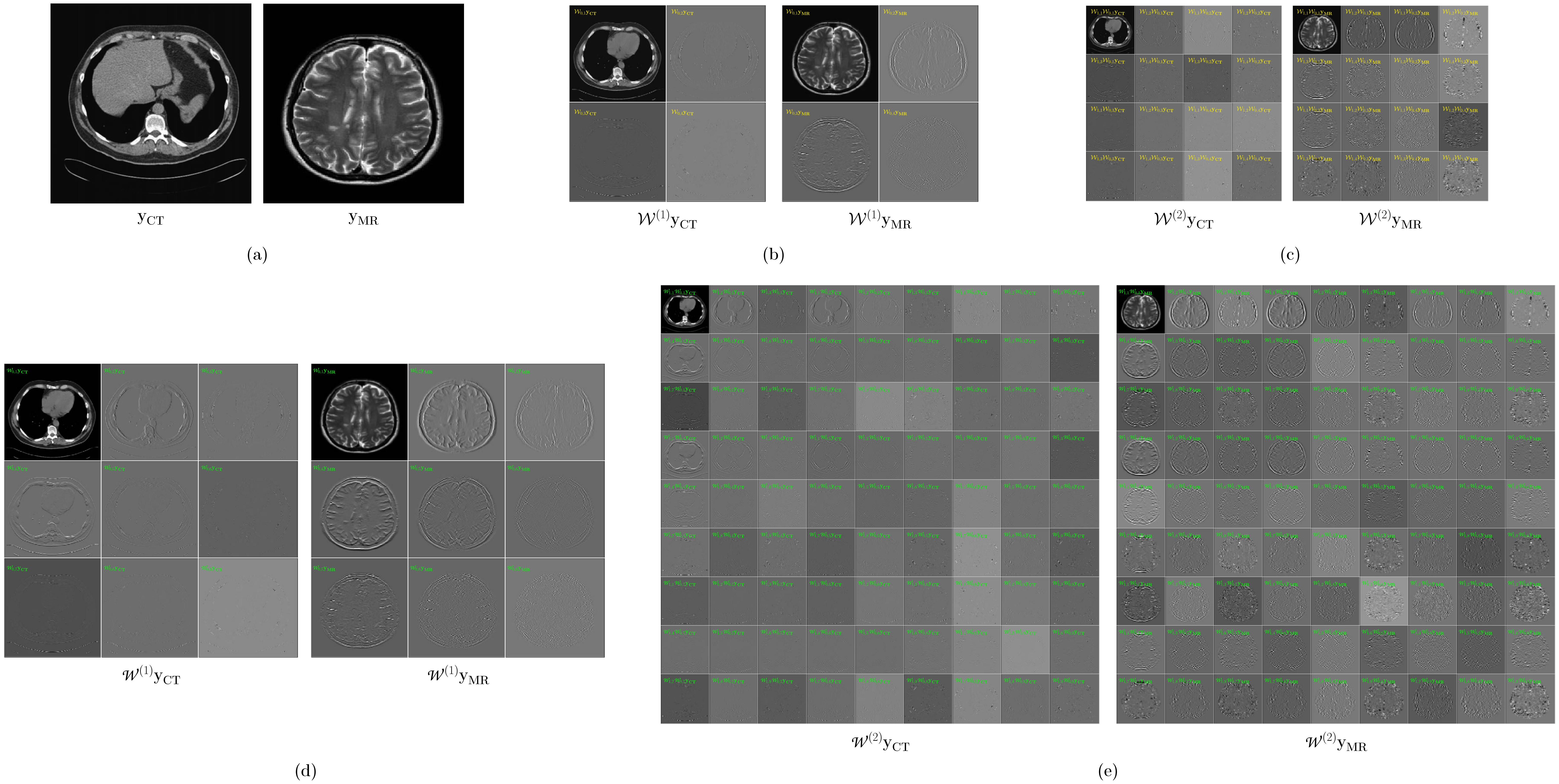}
	\caption{Framelet decomposition using various framelets(Db4 and B-spline); (a) Original CT image $\y_{\textbf{CT}}$ and MR image $\y_{\textbf{MR}}$, (b) - (c) first and second level framelet decomposition using Daubetchies 4 tab wavelet, denoted by $\mathcal{W}^{(1)}$ and $\mathcal{W}^{(2)}$, of the images $\y_{\textbf{CT}}$ and $\y_{\textbf{MR}}$, and (d) - (e) first and second level framelet decomposition using piecewise linear B-spline framelet, denoted by $\mathpzc{W}^{(1)}$ and $\mathpzc{W}^{(2)}$, of the images $\y_{\textbf{CT}}$ and $\y_{\textbf{MR}}$.}
	\label{FD}
\end{figure*}

Now, we define the first level framelet decomposition operator $\mathpzc{W}^{(1)}$ by
\begin{equation}
\mathpzc{W}^{(1)} = \begin{bmatrix} \mathpzc{W}_{0,0}^T, \mathpzc{W}_{0,1}^T, \cdots, \mathpzc{W}_{0,r}^T \end{bmatrix}^T
\end{equation}
where $\mathpzc{W}_{0,\alpha}$ is the $d^2/2^{-2} \times d^2$ matrix given by
\begin{align*}
\mathpzc{W}_{0,\alpha}\x = \downarrow(\x \circledast \textbf{q}_\alpha(-\cdot)), \forall \x \in \mathbb{R}^{d^2}.
\end{align*}
Here, $\downarrow$ stands for 2 dimensional down-sampling operator and $\circledast$ is convolution operator with stride 1. Likewise, we can define the second level framelet decompoosition $\mathpzc{W}^{(2)}$ by
\begin{align}
\nonumber \mathpzc{W}^{(2)}  = [ & (\mathpzc{W}_{1,0}\mathpzc{W}_{0,0})^T, \cdots, (\mathpzc{W}_{1,0}\mathpzc{W}_{0,r})^T, \\ &  \cdots, (\mathpzc{W}_{1,r}\mathpzc{W}_{0,0})^T, \cdots, (\mathpzc{W}_{1,r}\mathpzc{W}_{0,r})^T]^T
\end{align}
where $\mathpzc{W}_{1,\alpha}$ is the $d^2/2^{-4} \times d^2/2^{-2}$ matrix given by
\begin{align*}
\mathpzc{W}_{1,\alpha}\tilde{\x} = \downarrow(\tilde{\x} \circledast \textbf{q}_\alpha(-\cdot)), \forall \tilde{\x} \in \mathbb{R}^{d^2/2^{-2}}.
\end{align*}
We can continue the above process to define the $k$th level framelet decomposition operator $\mathpzc{W}^{(k)}$. Fig. \ref{FD} illustrates two examples of framelet decompositions using Daubechies wavelet(db4) \cite{Daubechies1988} and piecewise linear B-spline frame \cite{Shen2013}.

\begin{figure*}
	\centering
	\includegraphics[width=1\textwidth]{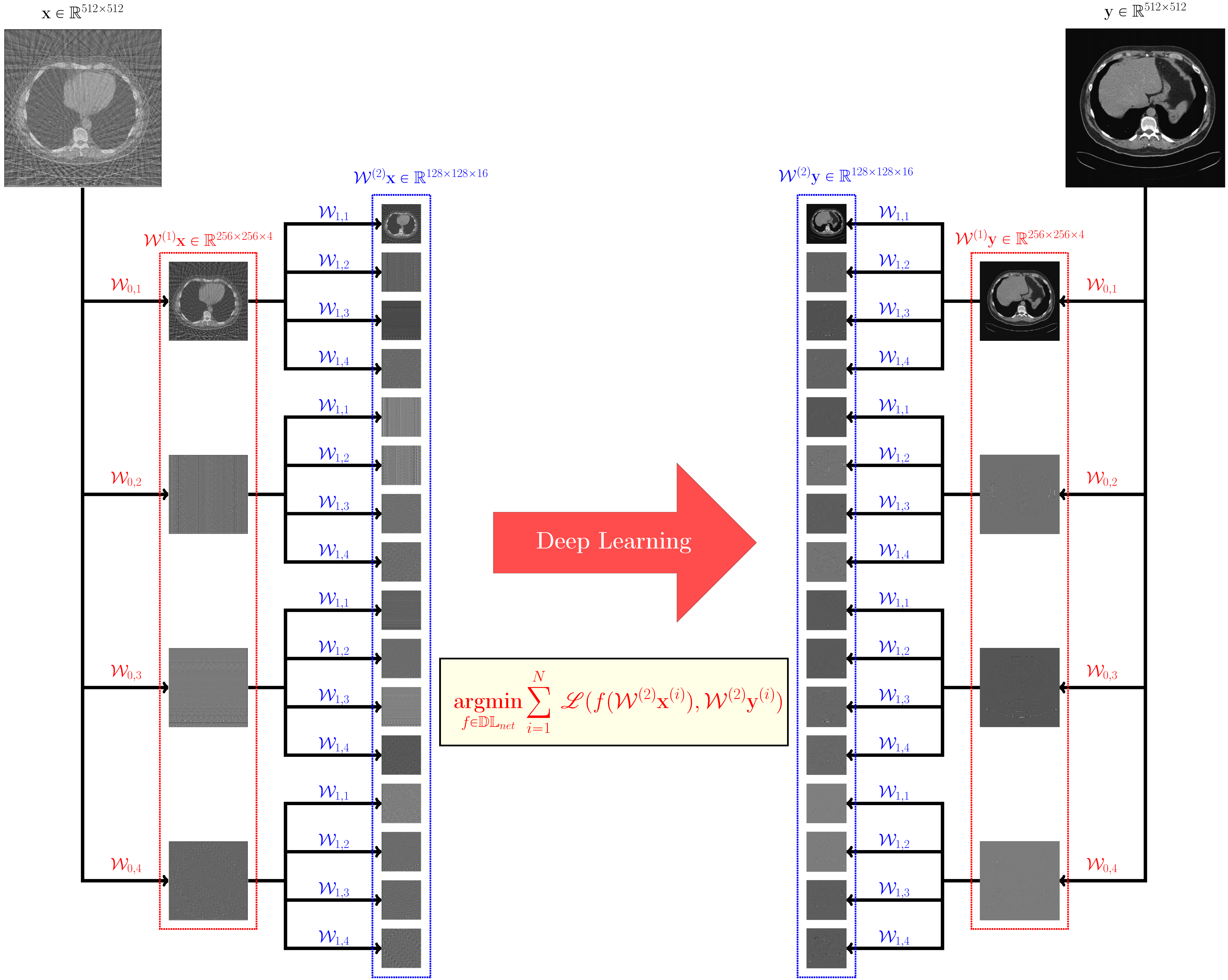}
	\caption{A one example of the proposed method with Daubechies 4 tab wavelet for sparse-view CT problem. The dataset $\{\x^{(i)},\y^{(i)} \}_{i=1}^{N}$ is decomposed by Daubechies 4 tap wavelet, $\mathcal{W}^{(2)}$. Our proposed network tries to infer the relation between  $\{\mathcal{W}^{(2)}\x^{(i)},\mathcal{W}^{(2)}\y^{(i)} \}_{i=1}^{N}$ }.
	\label{FADL}
\end{figure*}

Now, we are ready to explain our proposed deep learning network. Let $\mathpzc{W}$ and $\mathcal{W}$ be framelet decomposition operators. The proposed framelet pooling deep learning network aims to infer the relation between $\mathpzc{W}^{(k_1)}\x$ and $\mathcal{W}^{(k_2)}\y$ in the following least-squared minimization sense :
\begin{equation} \label{FADLN}
\textbf{f} = \underset{\textbf{f} \in \mathbb{D}\mathbb{L}_{\tiny \mbox{net}}}{\mbox{argmin}}\sum_{i=1}^{N}  \mathscr{L}  (\textbf{f}(\mathpzc{W}^{(k_1)}\x^{(i)}),\mathcal{W}^{(k_2)}\y^{(i)})
\end{equation}
Here, each $(\mathpzc{W}^{(k_1)}\x^{(i)})_{\alpha_1}$ and $(\mathcal{W}^{(k_2)}\y^{(i)})_{\alpha_2}$ are images with $d^2/2^{-2k_1}$ and $d^2/2^{-2k_2}$ pixel dimension, respectively. For example, let $\mathcal{W}^{(2)}$ be the second level Daubechies 4 tab wavelet decomposition. If the second level Daubechies 4 tab wavelet decomposition is taken for $\mathpzc{W}^{(k_1)}$ and $\mathcal{W}^{(k_2)}$ in the equation $\eqref{FADLN}$, the proposed deep learning method tries to find the function $\textbf{f}$ satisfying $\textbf{f}(\mathcal{W}^{(2)}\x)=\mathcal{W}^{(2)}\y$ in the sense of $\eqref{FADLN}$, as shown in Fig \ref{FADL}.

Compared to the direct deep learning scheme $\eqref{directDL}$, the framelet-pooling aided deep learning method $\eqref{FADLN}$ is expected to mitigate the total computational complexity and time caused by high dimensional data in the learning process. In this paper, we test only the case that training inputs and labels are decomposed using same framelet decomposition $\mathpzc{W}^{(k)}$. However, our method is not restricted only in this specific case.

\section{Experiments and Results} \label{Result}
\subsection{Experimental Set-up for Undersampled MRI}
Let $\{ \y_{\mbox{\tiny MR}}^{(i)} \in \mathbb{R}^{256 \times 256} \}_{i=1}^{N}$ denote the set of MR images reconstructed with the Nyquist sampling. Using $\{\y_{\mbox{\tiny MR}}^{(i)}\}$, we compute the training input $\{ \x_{\mbox{\tiny MR}}^{(i)} \in \mathbb{R}^{256 \times 256} \}_{i=1}^{N}$ by
\begin{equation}\label{OriginalNet1}
 \x_{\mbox{\tiny MR}}^{(i)} = \mathscr{F}^{-1}\mathcal{S}^* \underbrace{\mathcal{S}\mathscr{F}\y_{\mbox{\tiny MR}}^{(i)}}_{\mbox{\footnotesize $\textbf{P}^{\sharp}$}}
\end{equation}
where $\mathscr{F}$ is the 2D discrete Fourier transform, $\mathscr{F}^{-1}$ is the 2 dimensional discrete inverse Fourier transform, and $\mathcal{S}$ is a specifically user-chosen subsampling operator. In our experiments, we use the MR images $\y_{\mbox{\tiny MR}}^{(i)}$ obtained from T2-weighted turbo spin-echo pulse sequence with 4408 ms repetition time, 100 ms echo time, and 10.8 ms echo spacing \cite{Loizou2011}. The Fourier transform and its inverse are computed via \texttt{fft2} and \texttt{ifft2} in the Python package \texttt{numpy.fft}. Finally, for the sampling strategy, we choose the uniform subsampling with factor 4 and 12 additional low frequency sampling among total 256 lines \cite{Hyun2018}.

In order to test our proposed method, we decompose dataset using $k$ level framelet decomposition $\mathpzc{W}^{(k)}$ with various filter banks. We obtain
\begin{equation}\label{FrameNet1}
\{ \mathpzc{W}^{(k)}\x_{\mbox{\tiny MR}}^{(i)}, \mathpzc{W}^{(k)}\y_{\mbox{\tiny MR}}^{(i)} \}_{i=1}^{N}
\end{equation}
where both $\mathpzc{W}^{(k)}\x_{\mbox{\tiny MR}}^{(i)}$ and $\mathpzc{W}^{(k)}\y_{\mbox{\tiny MR}}^{(i)}$ contains $r^k$ pairs of $256/2^{2k} \times 256/2^{2k}$ image. Here, $k$ is the decomposition level and $r$ is the number of filter $\textbf{q}_{\alpha}$.

\subsection{Experimental Set-up for Sparse-view CT}
{Let $\{\y^{(i)}_{\mbox{\tiny CT}} \in \mathbb{R}^{512 \times 512}\}_{i=1}^{N}$ be a set of CT images reconstructed with the Nyquist sampling. The corresponding deep learning training inputs are computed in the following sense;
\begin{equation}\label{OriginalNet2}
 \x_{\mbox{\tiny CT}}^{(i)} = \mathscr{R}^{-1}\mathcal{S}^* \underbrace{\mathcal{S}\mathscr{R}\y_{\mbox{\tiny CT}}^{(i)}}_{\mbox{\footnotesize $\textbf{P}^{\sharp}$}}
\end{equation}
where $\mathscr{R}$ is the discrete Radon transform, $\mathscr{R}^{-1}$ is the filtered-back projection algorithm, and $\mathcal{S}$ is a user-defined sampling operator. In our implementations, we use the projection algorithm $\texttt{radon}$ and filtered back-projection algorithm $\texttt{iradon}$ in the Python package $\texttt{skimage.transform}$ for computing $\mathscr{R}$ and its inverse $\mathscr{R}^{-1}$ respectively. Uniform subsampling with factor $6$ in terms of projection-view is also used for $\mathcal{S}$ in \eqref{OriginalNet2}.

Applying the same process used to generate a dataset \eqref{FrameNet1} for undersampled MRI experiments, we obtain the following decomposed dataset for sparse-view CT problem;
\begin{equation}\label{FrameNet2}
\{ \mathpzc{W}^{(k)}\x_{\mbox{\tiny CT}}^{(i)}, \mathpzc{W}^{(k)}\y_{\mbox{\tiny CT}}^{(i)} \}_{i=1}^{N}
\end{equation}
where $\mathpzc{W}^{(k)}$ is a $k$ level framelet decomposition.

In our whole experiments, we use a first and second level framelet decomposition ($k=1,2$) with three different framelets (Haar wavelet(Haar), Daubechies 4 tap wavelet(Db4), and piecewise linear B-spline framelet(PL)).

\subsection{Network Configuration} \label{NetConf}
\begin{figure*}
	\centering
	\includegraphics[width=1\textwidth]{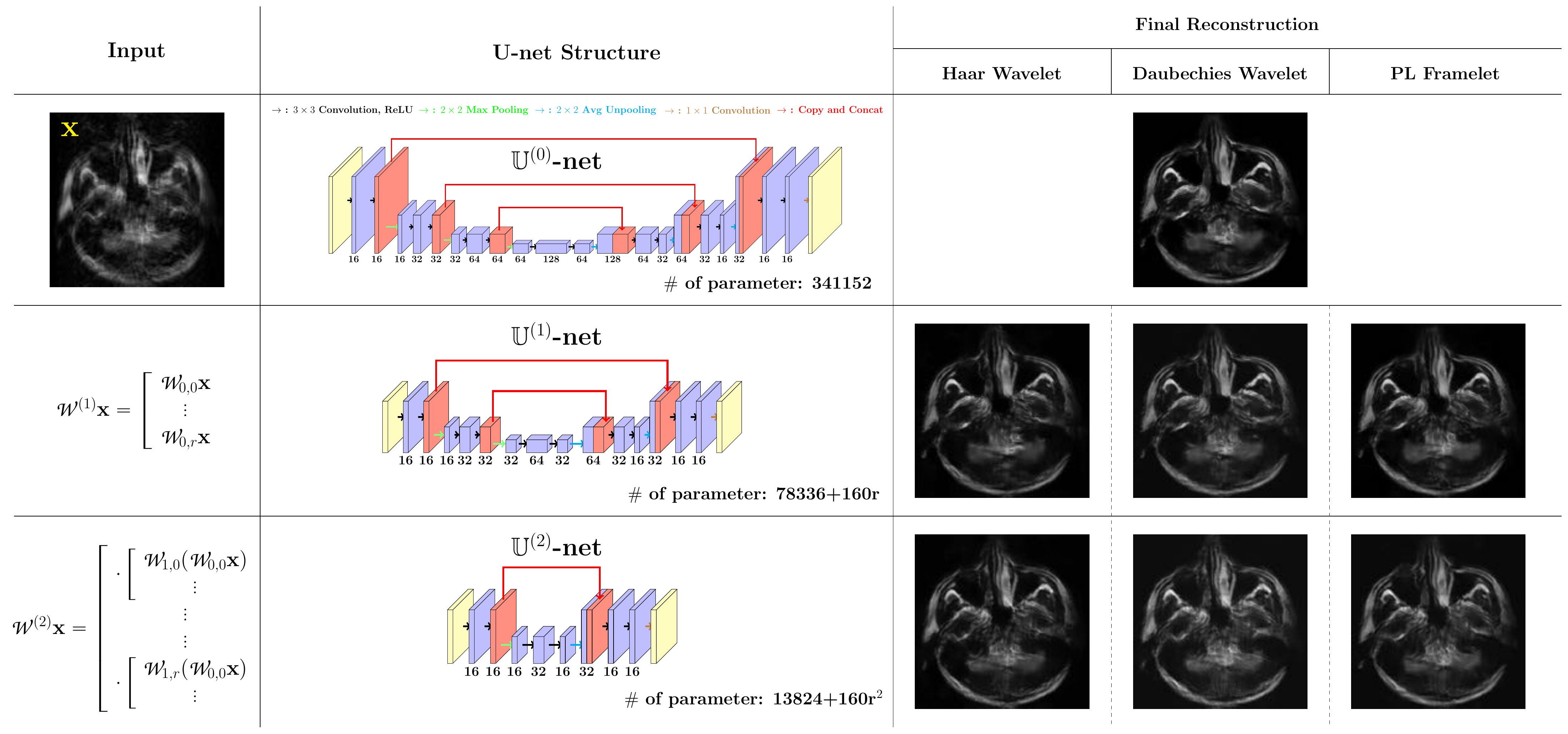}
	\caption{Image restoration performances of $\Bbb U^{(0)}\mbox{\footnotesize -NET}$ (unreduced), $\Bbb U^{(1)}\mbox{\footnotesize -NET}$ (one pooling), and $\Bbb U^{(2)}\mbox{\footnotesize -NET}$ (double pooling) using undersampled MRI and 1500 training data using various filter banks with $r$ filters. We set all networks to have 16 feature depth, which means that channels in the network paths are multiples of 16. Although the numbers of learning parameters and training speed are very different, the final results are very similar.}
	\label{Unet}
\end{figure*}

To test our proposed method, we adapt the U-net architecture \cite{Ronneberger2015}, as shown in Fig. \ref{Unet}, where the first half of network is the contracting path and the last half is the expansive path. At the first layer in U-net in Fig. \ref{Unet}, the input $\mathpzc{W}^{(k)}\x$ is convolved with the set of convolution filters $\textbf{C}^{(1)}$ so that it generates a set of feature maps $\mathbf{h}^{(1)}$, given by
$$
\mathbf{h}^{(1)} = \mbox{ReLU}(\textbf{C}^{(1)} \circledast_{1} \mathpzc{W}^{(k)}\x)s
$$
where $\mbox{ReLU}$ is the rectified linear unit $\mbox{ReLU}(x)=\max\{x,0\}$ and  $\circledast_{1}$ stands for the convolution with stride 1. We repeat this process to get $\mathbf{h}^{(2)}= \mbox{ReLU}(\textbf{C}^{(2)} \circledast_{1} \mathbf{h}^{(1)})$ and apply max pooling to get $\mathbf{h}^{(3)}$. Through this contracting path, we can obtain low dimensional feature maps by applying either convolution or max pooling. In the expansive path, we use the $2\times2$ average unpooling instead of max-pooling to restore the size of the output. To restore details in image, the upsampled output is concatenated with the correspondingly feature from the contracting path. At the last layer a 1$\times$1 convolution is used to combine each feature with one integrated feature \cite{Ronneberger2015}.

The U-net in the top row of Fig \ref{Unet} will be denoted by $\Bbb U^{(0)}\mbox{\footnotesize -NET}$. The U-net in the middle row, denoted by $\Bbb U^{(1)}\mbox{\footnotesize -NET}$, is the reduced network by eliminating two $3 \times 3$ convolution layers and one pooling/unpooling layer in the first and last part of $\Bbb U^{(0)}\mbox{\footnotesize -NET}$. Similarly, $\Bbb U^{(2)}\mbox{\footnotesize -NET}$ is the reduced network by eliminating $3 \times 3$ convolution layers and pooling/unpooling layer in the first and last part of $\Bbb U^{(1)}\mbox{\footnotesize -NET}$. Thus, this process can be viewed as the replacement of operations with unknown and trainable paramters into framelet operations with known and fixed paramters. In our experiments, $\Bbb U^{(0)}\mbox{\footnotesize -NET}$ is used to learn $f$ in the sense of direct learning \eqref{directDL}.  The reduced $\Bbb U^{(k)}\mbox{\footnotesize -NET}$  ($k=1,2$) is trained with $k$ level framelet decomposed dataset in the sense of \eqref{FADLN}.

\subsection{Experimental Result}
\begin{figure*}[h!]
	\centering
	\includegraphics[width=1\textwidth]{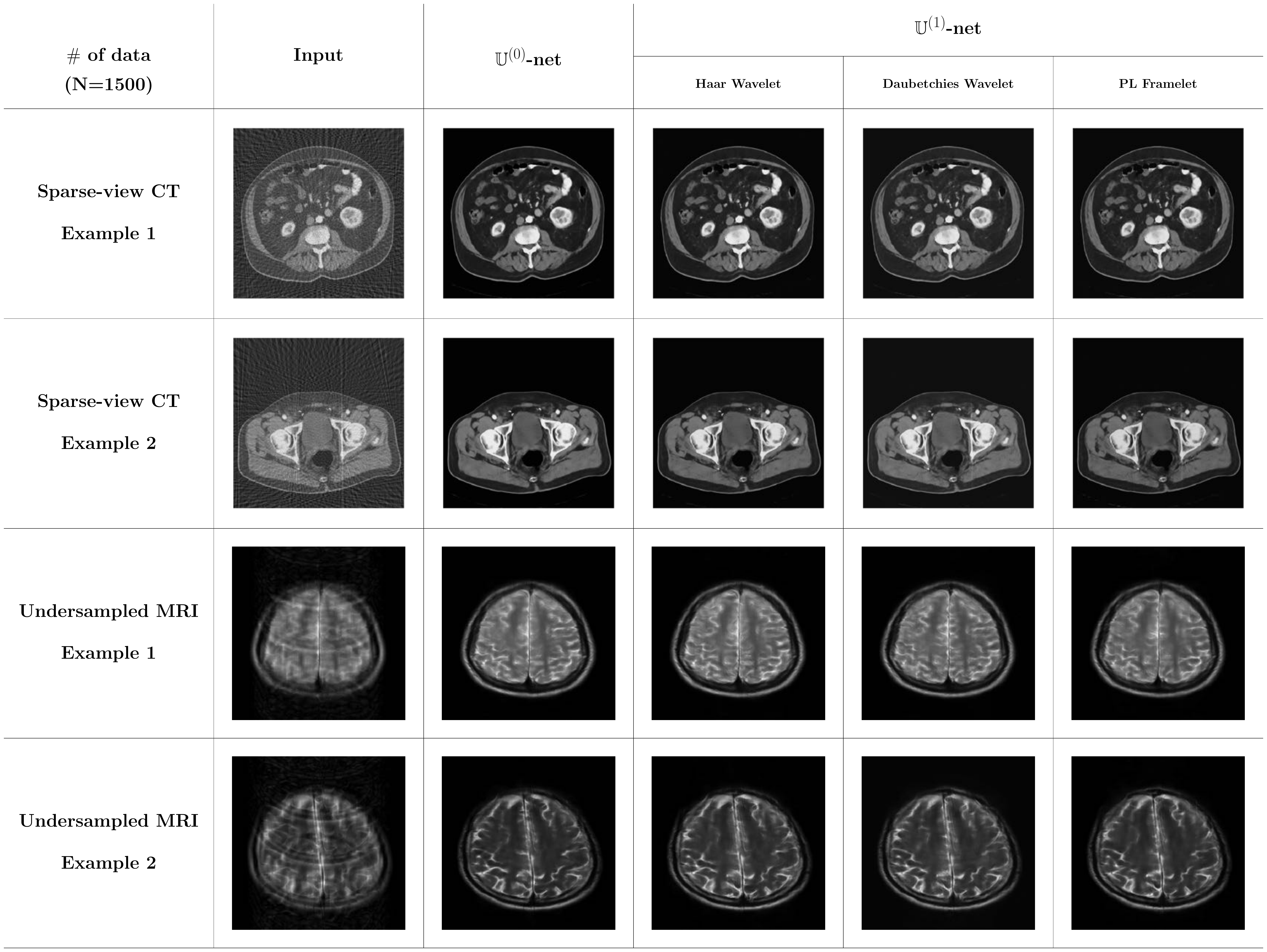}
	\caption{Image restoration performances of $\Bbb U^{(0)}\mbox{\footnotesize -NET}$ (unreduced), $\Bbb U^{(1)}\mbox{\footnotesize -NET}$ (one framelet pooling), and $\Bbb U^{(2)}\mbox{\footnotesize -NET}$ (double framelet pooling)  using undersampled MRI and 1500 training data. }
	\label{Unet2}
\end{figure*}
All training processes are performed in two Intel(R) Xeon(R) CPU E5-2630 v4, 2.20GHz, 128GB DDR4 RAM, and four NVIDIA GTX-1080ti computer system. We initialize all weights by a normal distribution with zero-centered and 0.01 standard deviation, under the Tensorflow environment \cite{Google}. We use the $\ell^2$ loss for the loss function $\mathscr{L}$. The loss function is minimized using the Adam Optimizer and the batch normalization for fast convergence \cite{Kingma2014,Ioffe2015}. For stability on training, the small learning rate $10^{-6}$ is used. In order to guarantee the convergence of loss function, the network is trained until the training loss seems to converge sufficiently.

\begin{table*}[h!]
	\centering
	\begin{tabular}[width=1\textwidth]{ c | c  | c | c | c | c | c | c }
	\multirow{2}{*}{$\#$ of training data} & \multirow{2}{*}{$\Bbb U^{(0)}\mbox{\footnotesize -NET}$} & \multicolumn{3}{c|}{$\Bbb U^{(1)}\mbox{\footnotesize -NET}$} & \multicolumn{3}{c}{$\Bbb U^{(2)}\mbox{\footnotesize -NET}$} \\ \cline{3-8}
	& & Haar & Db4 & PL & Haar & Db4 & PL \\ \hline
	 1500 &11.79803 &4.66432 &4.43048 &8.32320 &3.63646 &3.75758 & 12.98898 \\ \hline
		1000 &7.65065 &3.43165 &3.25852 &5.61659 &2.44117 &2.53221 & 8.312133 \\ \hline
		500 &3.76303 &1.70458 &1.75337 &2.65404 &1.27335 &1.31174 & 4.67596 \\ \hline
		100 &0.71829 &0.33526 &0.33408 &0.54698 &0.22354 &0.25795 & 0.94070 \\ \hline
	\end{tabular}
	\caption{Table of the average computational time per epoch in the undersampled MRI problem.}
	\label{table1}
\end{table*}

\begin{table*}[h!]
	\centering
	- MSE ($\sim 10^{-8}$) - \vspace{0.05cm} \\
	\begin{tabular}[width=1\textwidth]{ c | c  | c | c | c | c | c | c }
	\hline \hline
	\multirow{2}{*}{$\#$ of training data} & \multirow{2}{*}{$\Bbb U^{(0)}\mbox{\footnotesize -NET}$} & \multicolumn{3}{c|}{$\Bbb U^{(1)}\mbox{\footnotesize -NET}$} & \multicolumn{3}{c}{$\Bbb U^{(2)}\mbox{\footnotesize -NET}$} \\ \cline{3-8}
	& & Haar & Db4 & PL & Haar & Db4 & PL \\ \hline
		1500 &3.42766 &3.73420 &3.71604 &3.61357 &4.75285 &4.96253 & 4.72560 \\ \hline
		1000 &3.72376 &3.99806 &3.97593 &3.93189 &5.03000 &5.03232 & 4.91754 \\ \hline
		500 &4.41541 &4.47717 &4.62286 &4.37441 &5.38241 &5.47481 & 5.25260  \\ \hline
		100 &6.08867 &6.66704 &6.71441 &6.47448 &7.13172 &7.14136 & 6.91611 \\ \hline
	\end{tabular}
	\indent \vspace{0.05cm}
	\\ - PSNR -\\ \vspace{0.05cm}
	\begin{tabular}[width=1\textwidth]{ c | c  | c | c | c | c | c | c }
	\hline \hline
	\multirow{2}{*}{$\#$ of training data} & \multirow{2}{*}{$\Bbb U^{(0)}\mbox{\footnotesize -NET}$} & \multicolumn{3}{c|}{$\Bbb U^{(1)}\mbox{\footnotesize -NET}$} & \multicolumn{3}{c}{$\Bbb U^{(2)}\mbox{\footnotesize -NET}$} \\ \cline{3-8}
	& & Haar & Db4 & PL & Haar & Db4 & PL \\ \hline
		1500 &26.7023 &26.3271 &26.3447 &26.4676 &25.2337 &25.0426 & 25.2652 \\ \hline
		1000 &26.3346 &26.0158 &26.0411 &26.0942 &24.9815 &24.9722 & 25.0855 \\ \hline
		500 &25.5725 &25.5103 &25.3798 &25.6182 &24.672 &24.6044 & 24.7897 \\ \hline
		100 &24.1309 &23.7245 & 23.6861 &23.8651 &23.3834 &23.3959 & 23.5453 \\	\hline
	\end{tabular}
	\indent \vspace{0.05cm}
	\\ - SSIM - \vspace{0.05cm} \\
	\begin{tabular}[width=1\textwidth]{ c | c  | c | c | c | c | c | c }
	\hline \hline
	\multirow{2}{*}{$\#$ of training data} & \multirow{2}{*}{$\Bbb U^{(0)}\mbox{\footnotesize -NET}$} & \multicolumn{3}{c|}{$\Bbb U^{(1)}\mbox{\footnotesize -NET}$} & \multicolumn{3}{c}{$\Bbb U^{(2)}\mbox{\footnotesize -NET}$} \\ \cline{3-8}
	& & Haar & Db4 & PL & Haar & Db4 & PL \\ \hline
		1500 &0.81243 &0.79251 &0.79806 &0.79925 &0.74633 &0.73704 & 0.73648 \\ \hline
		1000 &0.80122 &0.78788 &0.78861 &0.79088 &0.73594 &0.73546 & 0.74440  \\ \hline
		500 &0.77455 &0.77260 &0.77219 &0.76996 &0.71341 &0.71464 & 0.71955 \\ \hline
		100 &0.70940 &0.69864 &0.70857 &0.72080 &0.64815 &0.64767 & 0.67378 \\ \hline
	\end{tabular}
	\caption{Quantitivative test error evaluations for undersampled MRI problem using three different metrics.}
	\label{table3}
\end{table*}

\newpage
Fig. \ref{Unet} and Fig. \ref{Unet2} show reconstruction results from $\Bbb U^{(0)}\mbox{\footnotesize -NET}$, $\Bbb U^{(1)}\mbox{\footnotesize -NET}$, and $\Bbb U^{(2)}\mbox{\footnotesize -NET}$. Three models show similar reconstruction performances, regardless of their originated problem and their original data dimension. Quantitative evaluations and comparisons for the application on the undersampled MRI problem are summerized in the Table \ref{table1} and \ref{table3}. For the sparse-view CT application, evaluations and comparisons are given in Table \ref{table2} and Table \ref{table4}. Table \ref{table1} and Table \ref{table2} shows comparisons of average computational time per epoch among $\Bbb U^{(0)}\mbox{\footnotesize -NET}$, $\Bbb U^{(1)}\mbox{\footnotesize -NET}$, and $\Bbb U^{(2)}\mbox{\footnotesize -NET}$. The average computational time is computed by dividing the total computational time by the total number of epoch. Table \ref{table3} and Table \ref{table4} contains test error evaluations and comparions using three different metrics; mean square error(MSE), peak signal to noise ratio(PSNR), and structure similarity(SSIM) \cite{wang2004}.

These experimental results support the fact that the proposed method reduces the total computational time efficiently and provides competitive results compared to the direct learning algorithm using high dimensional images. Namely, our reduced method provides very similar performance to the standard unreduced method ($\Bbb U^{(0)}\mbox{\footnotesize -NET}$), while reducing the computation time greatly by reducing the input dimension.

\begin{table*}[h!]
	\centering
	\begin{tabular}[width=1\textwidth]{ c | c  | c | c | c | c | c }
		\hline \hline
		\multirow{2}{*}{$\#$ of training data} & \multirow{2}{*}{$\Bbb U^{(0)}\mbox{\footnotesize -NET}$} & \multicolumn{3}{c|}{$\Bbb U^{(1)}\mbox{\footnotesize -NET}$} & \multicolumn{2}{c}{$\Bbb U^{(2)}\mbox{\footnotesize -NET}$} \\ \cline{3-7}
		& & Haar & Db4 & PL & Haar & Db4 \\ \hline
		1500 &39.47637 &17.29248 &18.34227 &31.42237 &12.09981 &11.75551  \\ \hline
		1000 &26.42527 &11.80329 &12.34189 &20.91642 &8.27648 &8.21492  \\ \hline
		500 &13.00667 &6.00089 &6.06389 &10.54422 &4.10279 &4.03506  \\ \hline
		100 &2.46522 &1.04074 &1.12465 &1.95324 &0.75294 &0.81197 \\
		\hline
	\end{tabular}
	\caption{Table of the average computational time per epoch in the sparse-view CT problem.}
	\label{table2}
\end{table*}
\begin{table*}[h!]
	
	\centering
	- MSE ($\sim 10^{-9}$) - \vspace{0.05cm} \\
	\begin{tabular}[width=1\textwidth]{ c | c  | c | c | c | c | c }
		\hline \hline
		\multirow{2}{*}{$\#$ of training data} & \multirow{2}{*}{$\Bbb U^{(0)}\mbox{\footnotesize -NET}$} & \multicolumn{3}{c|}{$\Bbb U^{(1)}\mbox{\footnotesize -NET}$} & \multicolumn{2}{c}{$\Bbb U^{(2)}\mbox{\footnotesize -NET}$} \\ \cline{3-7}
		& & Haar & Db4 & PL & Haar & Db4 \\ \hline
		1500 &2.62030 &2.92002 &2.95605 &2.93551 &3.87506 &4.28539  \\ \hline
		1000 &2.69498 &3.03115 &3.10120 &2.97395 &4.20075 &4.42756  \\ \hline
		500 &2.80345 &3.21323 &3.32694 &3.09397 &4.42089 &5.40228  \\ \hline
		100 &3.81432 &4.32176 &4.72559 &4.10138 &5.65774 &7.58266  \\
		\hline
	\end{tabular}
	\indent \vspace{0.05cm}
	\\ - PSNR - \vspace{0.05cm} \\
	\begin{tabular}[width=1\textwidth]{ c | c  | c | c | c | c | c }
		\hline \hline
		\multirow{2}{*}{$\#$ of training data} & \multirow{2}{*}{$\Bbb U^{(0)}\mbox{\footnotesize -NET}$} & \multicolumn{3}{c|}{$\Bbb U^{(1)}\mbox{\footnotesize -NET}$} & \multicolumn{2}{c}{$\Bbb U^{(2)}\mbox{\footnotesize -NET}$} \\ \cline{3-7}
		& & Haar & Db4 & PL & Haar & Db4 \\ \hline
		1500 &31.7860 &31.2960 &31.2357 &31.2723 &30.0371 &29.5805  \\ \hline
		1000 &31.6586 &31.1279 &31.0197 &31.2154 &29.6759 &29.4287  \\ \hline
		500 &31.4741 &30.8580 &30.6976 &31.0308 &29.4400 &28.5531  \\ \hline
		100 &30.0695 &29.5211 &29.1323 &29.7526 &28.3498 & 27.0967  \\
		\hline
	\end{tabular}
	\indent \vspace{0.05cm}
	\\ - SSIM - \vspace{0.05cm} \\
	\begin{tabular}[width=1\textwidth]{ c | c  | c | c | c | c | c }
		\hline \hline
		\multirow{2}{*}{$\#$ of training data} & \multirow{2}{*}{$\Bbb U^{(0)}\mbox{\footnotesize -NET}$} & \multicolumn{3}{c|}{$\Bbb U^{(1)}\mbox{\footnotesize -NET}$} & \multicolumn{2}{c}{$\Bbb U^{(2)}\mbox{\footnotesize -NET}$} \\ \cline{3-7}
		& & Haar & Db4 & PL & Haar & Db4 \\ \hline
		1500 &0.87631  &0.86937 &0.86660 &0.86959 &0.84795 &0.82865   \\ \hline
		1000 &0.87533 &0.86761 &0.86357 &0.86925 &0.84444 &0.82383   \\ \hline
		500 &0.87451 &0.86493 &0.85750 &0.86728 &0.83932 &0.80927   \\ \hline
		100 &0.86039 &0.84886 &0.84055 &0.84863 &0.82847 &0.80464   \\
		\hline
	\end{tabular}
	\caption{Quantitivative error evaluations for sparse-view CT problem using three different metrics.}
	\label{table4}
\end{table*}

We also test our proposed method with three different framelets and compare performances, as shown in Table \ref{table3} and \ref{table4} for the quantitative evaluation and Table \ref{table1} and \ref{table2} for the computational time. Experimental results report that Haar and Db4 Wavelet reduce the computational time more efficiently than PL framelet, but PL framelet exhibits the better performance than Haar and Db4 Wavelet. Compared to Haar and Db4 consisting of $4$ filter banks, PL framelet has $9$ filter banks (i.e. the number of filter banks equals the size of filters), which can increase the computational time. However, it should be noted that Haar and Db4 are orthonormal bases while PL framelet is a redundant tight frame system. This means that, thanks to the redundancy, it is likely that the error generated by the nonlinear deep learning process can lie in the nontrivial null space of the reconstruction operator, which can make the PL framelet yield better results than the orthonormal basis (Haar and Db4) \cite{bin2017}.
Lastly, we would like to mention that the computational time increases in the case of $\Bbb U^{(2)}\mbox{\footnotesize -NET}$ with PL framelet in the undersampled MRI problem, compared to the original network $\Bbb U^{(0)}\mbox{\footnotesize -NET}$. We can observe that the reduction of computational time depends on the feature depth of network. In order to reduce total computational complexities of experiments as possible, our networks are set to have 16 feature depth, as described in Fig \ref{Unet}. However, when the feature depth increases, $\Bbb U^{(2)}\mbox{\footnotesize -NET}$ with PL framelet also exhibits the computational time reduction ability, as shown in the Table \ref{table5}.

\begin{table}[h]
	\centering
	\begin{tabular}[width=1\textwidth]{ c | c | c | c  }
		\hline \hline
		 Feature depth & $\Bbb U^{(0)}\mbox{\footnotesize -NET}$ & $\Bbb U^{(1)}\mbox{\footnotesize -NET}$ & $\Bbb U^{(2)}\mbox{\footnotesize -NET}$ \\ \hline
		 16 & 7.65065 & 5.61659 & 8.312133 \\ \hline
		 32 & 12.78046 & 6.186188 & 8.634091 \\ \hline
		 64 & 26.20805 & 8.805362 & 8.979816 \\ \hline
	\end{tabular}
	\caption{Table of the average computational time per epoch in undersampled MRI problem, when using the proposed method with PL framelet and 1000 training data(N=1000).}
	\label{table5}
\end{table}

\section{Conclusion and Discussion} \label{Conclude}
In this paper, we proposed the framelet pooling aided deep learning network to reduce computational burdens in the training process. The proposed method decomposes large-scale learning tasks into several small-scale learning tasks through the framelet packet transformation so that we can handle large-scale medical imaging in a limited computing environment. Experimental results on undersampled MRI and sparse-view CT reconstruction problems show that our framelet pooling method is at least comparable to the standard deep learning based method, but is able to reduce total computational time in the training process significantly. Hence, we expect that our method is not limited to the two dimensional medical imaging problem. It seems possible that the framelet pooling method can be extended to deep learning problems with large-scale 3 dimensional medical imaging, which inevitably suffers from high computational complexity due to the high dimensionality of dataset.

In the experiments, we can see that the choice of filter banks indeed affects the performance of the proposed method. The use of tight frame can increase the reconstruction accuracy thanks to rich representation under the redundant system, but the computational time reduction ability can be marginal due to the increasing number of convolutions. In contrast, the orthogonal wavelet representation provides high computational time reduction by only using $4$ filters, but generates less accurate results. Hence, the future work will focus on the construction of framelet transformation which is both adaptive to a given task \cite{Gai2014} and computationally efficient. It would also be interesting to provide a theoretical analysis on the approximation property of our deep learning network.

\section*{Acknowledgments}
Hyun, Kim, Cho and Seo are supported by Samsung Science $\&$ Technology Foundation SSTF-BA1402-01, Hyun is supported by the National Research Foundation of Korea grants NRF-2018H1A2A1062505, Kim was supported by NRF grant NRF-2017R1E1A1A03070653.

\bibliographystyle{wileynum}

\end{document}